Designing a Safe Autonomous Artificial Intelligence Agent based on Human Self-Regulation

Mark Muraven

University at Albany

## Abstract

There is a growing focus on how to design safe artificial intelligent (AI) agents. As systems become more complex, poorly specified goals or control mechanisms may cause AI agents to engage in unwanted and harmful outcomes. Thus it is necessary to design AI agents that follow initial programming intentions as the program grows in complexity. How to specify these initial intentions has also been an obstacle to designing safe AI agents. Finally, there is a need for the AI agent to have redundant safety mechanisms to ensure that any programming errors do not cascade into major problems. Humans are autonomous intelligent agents that have avoided these problems and the present manuscript argues that by understanding human self-regulation and goal setting, we may be better able to design safe AI agents. Some general principles of human self-regulation are outlined and specific guidance for AI design is given.

Recently there has been some discussion on how to better align autonomous artificial intelligence actions with human interests. The issue, as discussed at length (e.g., Bostrom, 2014), is that an autonomous artificial intelligence agent, in blindly following its programming, may engage in actions that directly or indirectly hurt human welfare. Several potential solutions have been put forth, but to date, none are deemed sufficient or adequate to address the problem.  More broadly, there exists the general question of how to make an autonomous artificial intelligence agent work and what might be the key features of it. The goal of this manuscript is to argue that by understanding how humans autonomously reach their own goals, we can better design an autonomous artificial intelligence agent to both be autonomous and to pursue goals that align with human interests.

The underlying assumption behind this discussion is that humans have evolved and been socialized in ways that ensure optimal goal pursuit. Obviously, there are instances where humans' behavior is less than ideal, but as will be argued human goal behavior is a reasonable starting point for designing a self-regulation system for an autonomous agent. Thus, this manuscript provides a brief and introductory framework to research on human self-regulation.  There are many details and formalisms omitted for simplicity's sake. The point is to give a basic understanding of human self-regulation that may help in the design of an autonomous artificial intelligence agent as well.

In particular, understanding human self-regulation may guide research on artificial intelligence because humans seem to avoid many of the hypothesized pitfalls that an artificial intelligence may face. For instance, the obsessive pursuit of one final goal, analogous to the paperclip optimization problem, is rare in humans. Humans do not single-mindedly pursue one goal to the exclusion of all others. Foreshadowing some of the dynamics of the model, humans that pursue a goal beyond reasonable balance are considered mentally ill. Addicts, people suffering from obsession and compulsions, and those who engage in self-damage suffer from understandable and avoidable problems with their self-regulation system. Indeed, treatment for those conditions usually addresses how to better self-regulate. Thus, in humans, a correctly functioning self-regulatory system is needed to maintain mental health. Moreover, humans' ability to self-regulate is very robust as most people can adapt to highly dynamic environments and remain focused on long-term goal pursuit even when unexpected obstacles arise. These are traits that would be highly valued in an autonomous artificial intelligence agent. As described below, there are clear features of humans' goal system that helps to keep it stable and balanced that may be easily emulated in an artificial intelligence agent.

The other benefit from understanding how the human goal system operates is that most people engage in behaviors that they believe will benefit themselves and other people. Obviously, there is some level of self-deception and flexibility in definition (the dictator who puts opponents in jail may convince himself that it is for the good of society), but theories of human self-regulation suggest that there are checks and balances on such self-deception. These checks built into the self-regulation system help to explain why most people follow socially approved goals in socially appropriate ways. Understanding human self-regulation may therefore help in designing autonomous artificial intelligent agents that avoid undesired and unconventional solutions to problems (e.g., making everyone happy by putting wires into the brain's pleasure centers).

## Feedback Loops

Self-regulation is a broad and encompassing topic within modern cognitive science. At its heart, it describes how individuals select purposive goals, stay on track toward those goals, and correct for changes in the environment as they move toward their goals. The central idea is that there is a feedback control process that helps people reach their goals. The basic framework for understanding human self-regulation follows from cybernetic models of feedback control (Carver & Scheier, 1981) that suggests people compare their current state to some goal, standard or ideal. That is, the feedback system compares incoming information to a reference value. If a discrepancy is noted between the current state and ideal state, some action takes place to try to minimize (or maximize, if the goal is to be unlike something) that difference. Thus, at its core there are four simple parts to a feedback loop: current state monitoring, ideal state monitoring, a comparator system to evaluate differences between those two states, and an output function to reduce the discrepancies between those two states.

The critical feature of cybernetic control requires the agent to have some idea of what actions may reduce (or enlarge) the discrepancy between the current state and goal state. Learning and training helps make this self-regulation process more efficient so the agent can quickly decide what actions are likely to lead to the desired outcome (Taylor, Pham, Rivkin, & Armor, 1998). However, as explained below, the autonomous agent should be open to feedback and is constantly self-monitoring, so when an action does not lead to the desired outcome, adjustments can be made.

Complex and autonomous behavior can begin to emerge when multiple conflicting feedback loops are nested within a hierarchical structure. The critical feature is not the cybernetic feedback loop itself, but rather the way multiple conflicting loops interact. It is through these chaotic and dynamic interactions that autonomy emerges. Moreover, these interactions among loops are the critical feature in maintaining goal focus in changing environments and avoiding actions that would damage themselves and others.

## Basic Features

### Hierarchical Goals

One critical feature of the feedback loop is that goals are arranged in an hierarchical structure (Elliot, 2006). Every goal has an underlying motivation (superordinate goal) and activates sub-goals. For example, a person who wishes to make a friend (goal) may be doing that to fulfill the need to feel connected to others (superordinate goal). They might pursue that goal by texting someone they met (sub-goal). Progress on each of these goals at every level of the hierarchy is monitored and controlled by the cybernetic feedback loops previously described. Lower level goals are more concrete and discrete, whereas higher level goals tend to be more abstract.

The top-most goals should be agent-defining and represent the fundamental values of the autonomous agent, although they are not often brought to conscious awareness. These top level goals are fundamental and basic. That is, they exist in everyone, cannot be ignored, and are critical to the successful functioning of the autonomous agent. For humans, these top-level goals are likely the product of evolution and thus innate (Deci & Ryan, 2000). There may be individual differences in how strongly each goal is pursued, but they are valued to some extent in every person. Even in cases of disordered behavior, such as addiction, a mixture of these goals is still active (Graham, Young, Valach, & Alan Wood, 2008). These top level goals are basic to human nature and ultimately drive all behavior.

For humans, the exact enumeration of these top-level goals is a matter of some controversy, but most accounts include a need to feel competent, a desire to have a meaningful impact on the world, a dislike of inconsistencies as well as more basic concerns such as a need for survival and social connections. For the purpose of an autonomous artificial intelligence agent, an exact and complete description of these top-level goals may be less important than the fact that there are multiple conflicting goals that are aligned with human desires.

Given the hierarchical nature of goals, each of these top level goals can be fulfilled in different and perhaps competing ways. For example, the goal of belongingness or relatedness may lead to the goal of calling friends more often, becoming more active in a social movement, and showing more affection to one's family.  These lower level goals also may conflict so that progress on one goal may hinder progress on another goal. People's habitual solutions to these conflicts is an important source of individual differences.

## Conflicting Goals

Given that people have multiple goals active at the same time, these goals may often be in conflict. The pursuit of one goal may block, interfere or even undo the pursuit of another. For example, the desire to feel competent may lead to risk taking, which conflicts with the need for survival. The interfering nature of these goals is a critical feature of a successful autonomous agent and necessary for proper functioning.

If an autonomous agent has to balance multiple, conflicting goals, there has to be a prioritization process to resolve these competing desires (Stroebe, Mensink, Aarts, Schut, & Kruglanski, 2008). Given the wide range of potential goals, the autonomous agent needs some way to narrow the choices. Some general principles seem to be at work. People often try to find situations that will maximize their progress to as many goals as possible (multifinality; Förster, Liberman, & Friedman, 2007; Kruglanski et al., 2013). For example, people like to study in groups, so both social and competency goals can be simultaneously filled. They also seek goals that may be met in different ways, thus allowing for multiple paths to the desired endpoint (equifinality; Kruglanski, Pierro, & Sheveland, 2011).

In general, if progress toward a goal is blocked, an autonomous agent typically should try to find other ways to fulfill that goal. Just because current progress is frustrated, the goal should not be dropped. Instead, the agent should search through alternative paths to find another way to reach the goal. As a general principle, the higher the goal is in the hierarchy, the less willing the autonomous agent should be to drop it. If a call to a friend goes unanswered, the person will likely try another friend before giving up on fulfilling a desire to fulfill a need for social connection. Even when blocked multiple ways, people will seek other ways to fulfill the same goal. If no friends answer their phone, a person may start posting on Facebook to gain some sense of social connection.

More generally, the fields of psychology, economics, decision making, and sociology have put forth several different but ultimately related theories of motivation and goal selection.  Recently, Steel and König (2006) integrated these theories into their temporal motivation theory (TMT), which is based on four core features: value, expectancy, time, and loss versus gains. Value is determined by anticipated satisfaction of goal completion, whereas expectancy is the perceived probability that that outcome will occur. Temporal discounting results in events closer in time being more influential. Finally, gains and losses are weighted differently, so that losses hurt more than gains and the same outcome could be perceived as a gain or loss based on the context.  This theory can be integrated with theories that limit the search space for potential alternatives, like Decision Field Theory (Busemeyer & Diederich, 2002) to further increase its utility. This model likely provides a strong basis for understanding goal selection by an autonomous agent.

## Goals Reprioritization

Although the TMT model does account for temporal discounting, it is still a relatively static perspective on goal pursuit. It fails to account for changes over time and interactions between multiple, conflicting goals. An autonomous agent should dynamically update its goals as the situation changes and as progress is made. If a person wants to build a shed and unexpectedly runs out of nails, the goal to buy more nails gets elevated in priority. If the store is closed, another goal may become active, such as paint the trim or respond to unanswered e-mails. For instance, research has found that goals for examinations changed as students' perception of their course grade changed (Lord & Hanges, 1987). Lower-level goals are subject to revision as circumstances change, based on feedback from more abstract, higher-level goals. Thus, goals must be allowed to change as the situation changes.

The dynamics of the self-regulation system must be changeable as well. Some goals should be hewed to closely and resist change. Other times the system should be more open and flexible. For instance, in times of exploration, creativity or difficulty, the system may look to internal and external cues to be very open to new goals. Other times, the autonomous agent should ignore distractions or other goals in order to maximize pursuit of the currently active goal.

Goal dynamics should depend on both bottom up processing based on the immediate value of the goal and top down processing, which considers the overall implication of the goal for the autonomous agent. Bottom-up models, like the multiple-goal pursuit model (MGPM; Ballard, Yeo, Loft, Vancouver, & Neal, 2016) dynamically update the discrepancy between current and goal state, as well as the available and required resources needed to reach the goal, which determines the motivation to act on that goal at any moment in time.  This means there should be regular evaluations of progress and alternate goals. The amount of processing dedicated to this reevaluation should probably depend on the nature of the goal and its place in the hierarchy of goals. Moreover, a successful autonomous agent should learn from prior experience and anticipate the effects of its behaviors, so that goals are dynamically updated based on past actions. On the other hand, top-down models of goal dynamics also need to be considered (Lynn, Wormwood, Barrett, & Quigley, 2015). Such models assume that decisions are made within the larger context of the decision maker's life. Thus prior history and future desires are weighed in the decision-making process.

## Feedback

Implicit in all these models is a need for regular assessment of goal progress. A successful autonomous agent must get regular feedback on its progress toward its goals. In humans, there is good evidence to suggest that emotions serve this purpose. Researchers have theorized that people have a meta-monitoring process that focuses on their rate of progress toward goals. Goals associated with positive affect are more likely to be prioritized (Custers & Aarts, 2007). Likewise, the more a behavior matches critical goals, the more positive people feel (Chirkov, Ryan, Kim, & Kaplan, 2003). When progress is perceived to be adequate, people experience positive affect; inadequate progress results in negative affect. This emotional feedback process allows people to track their progress toward multiple goals and hold models of desired progress toward these goals. Based on this feedback, people will prioritize some goals over others and increase or decrease effort on those goals.

Other self-conscious emotions, such as shame, guilt, and pride may provide additional goal related feedback (Tangney, 2003). These emotions are powerful guides to behaviors and strongly motivate self-regulation. Additional emotional information includes intensity and arousal level, based on the type of goal being pursued and changes in rate in progress. Each level in the hierarchy in goals is likely separately monitored, although feelings associated with low level goals may be more fleeting than higher level goals.

Further complicating matters, the feedback from progress may lag behind actual output. That is, it may take some time for the effects of the behavior to become evident. Thus, an autonomous agent should be able to reward itself from small actions that help move it toward goals. Such intrinsic motivation helps to ensure that goals are not abandoned when feedback is long in coming (Deci & Ryan, 1991).

The criterion for the feedback is likely multi-dimensional, especially for more abstract goals. For example, to determine whether the autonomous agent is making a positive change in the world (competence goal), the agent may look to others' reaction, changes in the environment, and degree of challenge felt during the activity (Fisher, 1978). There should be some flexibility in the criterion as well—new situations should lead to looser criteria than familiar situations. Experience should add more criteria to evaluate success and repeated success or failures should lead to recalibration of the criteria so the agent's affective experiences remains relatively stable.

In short, any autonomous agent needs a method to monitor progress toward its short and long-term goals. This feedback monitoring process seems to be largely built into human neurological system. Because of its fundamental and critical importance, it seems to be universal in mentally healthy people and resistant to change. Obviously, people may seek ways to modify their emotional state outside of goal pursuit, but society generally frowns upon long-term modulation of the emotional-feedback system (e.g., through drug use). Thus, an autonomous artificial intelligence agent will need a clear and dynamic feedback system that is difficult to ignore, resistant to change and not easily substituted by other experiences.

## Goal Fatigue

Finally, an autonomous agent needs to avoid getting stuck on a goal, even when that goal is seen as important and progress is being made. As explained above, autonomous agents should have multiple goals they wish to fulfill at any time but it is unlikely that they can fulfill more than one or two with any action. Thus, the agent needs some way to balance these conflicting and interfering goals.

Because making progress on one goal often means forestalling progress on another, the autonomous agent needs to have ways to select one goal and suppress all others. This ensures that the agent stays on task and does not get distracted by temporary or fleeting desires that might be triggered by the environment. Once a goal is selected, the agent should make certain that adequate resources are devoted to the task to secure its completion. Indeed, humans who are poorer at suppressing competing desires and staying on task suffer numerous problems in school, work and life (Tangney, Baumeister, & Boone, 2004). Research has shown that non-prioritized goals are shielded from interferences, so that these unselected goals are suppressed and less likely to be remembered (Shah, Friedman, & Kruglanski, 2002).

Choosing to prioritize one goal (or set of goals) over other goals should only be temporarily acceptable, however. The autonomous agent should have a built in fatigue process, so that the agent can be responsive to environmental change. Given the dynamics of goal pursuit, as the environment changes, new goals should be selected. This ensures that the agent can pursue complex goals in changing situations, and even account and correct for its own actions.

More profoundly, the agent should not allow any goal to be suppressed indefinitely even when the environment is static. The longer other goals are suppressed, the weaker that suppression should become. The autonomous agent should be less able to hold back competing goals and more responsive to environmental cues. This will lead to increased chances of goal switching. This is different from simple hyperbolic discounting, because it applies even to pleasurable activities and when progress toward the goal is acceptable. The agent may switch from higher priority to lower priority goals simply because in pursuit of the higher priority goal, the lower priority goals had to be suppressed.

There is extensive evidence in humans that self-control comes at a cost and is limited (Muraven, 2012; Muraven & Baumeister, 2000). Suppressing competing urges weakens inhibition, so that the ability to suppress alternative goals becomes weaker. Eventually, current goal pursuit is dropped and an alternative goal is pursued. Thus, no goal will be pursued indefinitely; people have an inbuilt mechanism that causes their ability to suppress alternatives to fail.

Although this failure of self-control is a problem for many people, it clearly has benefits as well. It prevents people from obsessively pursuing any goal, even if that goal is pleasurable and desired. People stop working, creating art, playing video games or even taking heroin, at least for a little while to pursue different goals. Thus, the failure of self-control in humans, although associated with negative consequences, may be critical to maintaining successful autonomous self-regulation. This failure of inhibition prevents autonomous agents from obsessive pursuit of any goal and helps ensure that behavior is balanced between goals.

This fatigue process may itself be open to feedback. For example, under certain conditions, the inhibition process may itself be strengthened or weakened to help the agent remain on task or allow itself to be distracted by higher priority tasks. Thus, things like external rewards or importance of the goal may allow the agent to at least temporarily ignore the fatigue feedback and continue working.

## Conclusion

Although competing goals seem like a feature to avoid in an autonomous agent, having conflicting goals is actually critical to the long-term stability of the agent. Such interfering goals prevent autonomous agents from pursuing one goal to the exclusion of all others, thereby helping to avoid situations of unwanted persistence, like the paperclipper maximizer problem.

Moreover, by having multiple goals active at once, problems with specifying one general law to guide an autonomous agent are reduced. Instead, the autonomous agent will seek an unstable equilibrium between several goal states, which prevents any one goal becoming predominant. The flexible feedback system further pushes the agent to avoid extremes, while permitting a level of situational adjustment and self-correction.

The fact that goals are arranged in hierarchies also helps in the design of an autonomous artificial intelligence. Only the top level goals need to be specified; the lower level goals that fulfill these goals can be discovered through exploration. The flexible feedback system helps to ensure that as the agent develops more complexity it will remain consistent with initial programming goals. That is, the same top-level goals that are active in babies (e.g., make social connections, gain competence) are active in adults, only they are pursued in a more complex way.

For the creation of an autonomous agent, specification of general, top-level groups that generally serve the welfare and betterment of humanity should help ensure alignment of the agent with human goals. In humans, these goals have been selected by evolution and society to help guide human behavior and thus have been tested as successful. The rare human that deviates or fails to balance these goals will undergo treatment to try to restore harmony and retraining to select appropriate goals, thus suggesting that human misery is not a product of these goals but rather a failure of self-regulation itself. An autonomous agent with a strong and well-designed self-regulatory system may therefore avoid these problems.

That is not to say that human self-regulation is completely understood. For instance, the nature of the topmost goals needs some level of specificity. Research on humans has identified many likely candidates, although there remains some disagreement. There also needs to be some work on how to formally specify these goals in order to assess progress. However, this is not an intractable problem, because as noted above, each level of goal spawns multiple, competing lower level goals. Given that each of these goals likely interfere and that the ability to suppress all other goals weakens over time, a misspecification in a goal will be corrected in the long run.

In summary, a consideration of human self-regulation may help in the design of a more general autonomous agent. Key features of human self-regulation include a hierarchical goal structure, multiple conflicting goals, an emotional feedback system and goal fatigue. This self-regulation system relies on relatively simple rules that generate an unstable equilibrium and thus may be ported to an artificial intelligence system. Moreover, there is reason to believe that an autonomous agent based on these principles will more likely be aligned to human goals and avoid poorly optimized, nonsensical, or monomaniac goal pursuit.